\documentclass{article}

\usepackage[usenames,dvipsnames]{xcolor}
\usepackage{graphicx}
\usepackage[frozencache=true,cachedir=minted-cache]{minted}

\usepackage{subfigure}
\usepackage{booktabs} % for professional tables

\usepackage{algorithm}
\usepackage[algo2e]{algorithm2e}

\usepackage{amsmath}
\usepackage{amsthm}
\usepackage{tabularx}
\theoremstyle{definition}
\newtheorem{definition}{Definition}[section]

\usepackage{hyperref}
\usepackage[accepted]{mlsys2024}

\usepackage{mlsys2024}
 \usepackage{listings}
 \usepackage[framemethod=TikZ]{mdframed}
\newmdenv[backgroundcolor=gray!20]{graybox}

\mlsystitlerunning{CompCodeVet}

\begin{document}

\twocolumn[
\mlsystitle{CompCodeVet: A Compiler-guided Validation and Enhancement Approach for Code Dataset}

% unless you've provided the [accepted] option to the mlsys2024


\mlsyssetsymbol{equal}{*}

\begin{mlsysauthorlist}
\mlsysauthor{Le Chen}{isu}
\mlsysauthor{Arijit Bhattacharjee}{isu}
\mlsysauthor{Nesreen K. Ahmed}{intel}
\mlsysauthor{Niranjan Hasabnis}{intel}
\mlsysauthor{Gal Oren}{technion}
\mlsysauthor{Bin Lei}{uconno}
\mlsysauthor{Ali Jannesari}{isu}
\end{mlsysauthorlist}

\mlsysaffiliation{isu}{Department of Computer Science, Iowa State University, Ames, USA}
\mlsysaffiliation{intel}{Intel Labs, Santa Clara, CA, USA}
\mlsysaffiliation{technion}{Computer Science Department, Technion - Israel Institute of Technology, Haifa, Israel}
\mlsysaffiliation{uconno}{Department of Computer Science & Engineering, University of Connecticut,  Storrs, USA}

\mlsyscorrespondingauthor{Le Chen}{lechen@iastate.edu}
\mlsyscorrespondingauthor{Ali Jannesari}{Jannesar@iastate.edu}

\mlsyskeywords{Machine Learning, MLSys}

\vskip 0.3in

\begin{abstract}
Large language models (LLMs) have become increasingly prominent in academia and industry due to their remarkable performance in diverse applications. As these models evolve with increasing parameters, they excel in tasks like sentiment analysis and machine translation. However, even models with billions of parameters face challenges in tasks demanding multi-step reasoning.
Code generation and comprehension, especially in C and C++, emerge as significant challenges. While LLMs trained on code datasets demonstrate competence in many tasks, they struggle with rectifying non-compilable C and C++ code. Our investigation attributes this subpar performance to two primary factors: the quality of the training dataset and the inherent complexity of the problem, which demands intricate reasoning.
Existing "Chain of Thought" (CoT) prompting techniques aim to enhance multi-step reasoning. This approach, however, retains the limitations associated with the latent drawbacks of LLMs.
In this work, we propose CompCodeVet, a compiler-guided CoT approach to produce compilable code from non-compilable ones. Diverging from the conventional approach of utilizing larger LLMs, we employ compilers as a teacher to establish a more robust zero-shot thought process. The evaluation of CompCodeVet on two open-source code datasets shows that CompCodeVet can improve the training dataset quality for LLMs.

% Among these tasks, code generation and comprehension stand out as particularly noteworthy. 
% Due to the importance of compilability of C and C++ programs, this work targets the compilable C and C++ code generation. Our analysis reveals that even LLMs specifically trained with codes are struggling with fixing non-compilable C and C++ code to be compilable. 

%  We extract 100k compilable C and C++ code from these two datasets and create a compiler-guided CoT for compilable code generation.

% While numerous LLMs have been trained with code, their proficiency in comprehending high-performance computing (HPC) code remains an evolving area of research.
% A model's performance is inextricably linked to the quality of its training data. Most current LLMs are trained or fine-tuned using publicly available code from platforms like GitHub and Stack Overflow. However, the quality of these datasets has not been extensively assessed. In this work, we propose CompCodeVet where we critically evaluate these code datasets, identifying and rectifying common errors like incomplete code snippets and mislabeled programming languages. Furthermore, we verify the compilation status of the code samples and employ a fine-tuned model to generate compilable code versions.
\end{abstract}
]



% \printAffiliationsAndNotice{\mlsysEqualContribution} % otherwise use the standard text.

\section{Introduction}
\label{sec: intro}
In recent years, the field of artificial intelligence has witnessed the meteoric rise of large language models (LLMs)\cite{NEURIPS2020_1457c0d6}\cite{chen2021evaluating}\cite{chowdhery2022palm}\cite{openai2023gpt4}. These models, built on colossal amounts of data and high computational capacities, have shown impressive results in a multitude of natural language processing tasks. Their capacity to understand, generate, and even exhibit creativity in language processing tasks has made them a focal point in machine learning research and applications.

The journey of machine learning in the realm of code analysis is not nascent. Researchers and developers have long worked on automating tasks such as code analysis, bug fixing, and code optimization. With the emergence and subsequent dominance of LLMs in the AI landscape, there has been a renewed and increasing interest in harnessing their power specifically for code-based tasks. This includes code analysis, code completion, and even automated bug detection.

\begin{figure}[!ht]
  \begin{center}
     \includegraphics[width=0.4\textwidth]{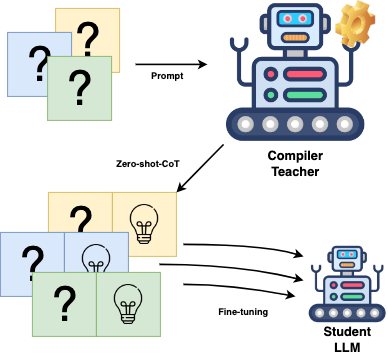}
     \caption{CompCodeVet: Diverging from the conventional approach of utilizing larger LLMs, we employ compilers (represented by the gear in the robot's hand) as teachers to establish a more robust zero-shot thought process. A smaller LLM is fine-tuned for compilable code generation.}
    \label{fig: intro}
  \end{center}
  % \vspace{-5mm}
\end{figure}

For effective training, LLMs demand voluminous amounts of data. While there exists a plethora of resources for natural language texts, the same cannot be said for code data. Presently, data for training LLMs on code-based tasks is primarily gathered from two major sources. One being, open-source repositories such as GitHub which offer a treasure trove of code from diverse projects and languages. 
The other are code snippets embedded within website content, often seen in platforms like StackOverflow where users share and discuss code snippets for problem-solving. However, relying on these sources presents the followin challenges:

\begin{enumerate}
    \item Incomplete code data: A significant portion of the code from these sources, particularly from online discussions, might be fragments rather than complete code. This incomplete nature can hinder the efficacy of LLMs when tasked with code auto-completions.
    \item Inaccurate file extensions and mislabeling: Code snippets extracted from inline website content frequently lack associated file extensions. This, combined with occasional mislabeling of the programming language used, can impede tasks like code classification.
    \item Low compilability: A non-trivial amount of code from these resources may not be directly compilable or might have inherent errors. This could compromise the quality of code generated by LLMs, as they might learn from flawed or erroneous code patterns.
\end{enumerate}

Recognizing these challenges, this paper sets forth a robust framework called CodeCompVet. The primary aim is to meticulously examine the current code datasets employed for training LLMs and subsequently enhance their quality and completeness. By addressing the inherent flaws and improving the quality of the training datasets, we aspire to push the boundaries of what LLMs can achieve in the domain of code analysis and generation while using a smaller number of parameters which in turn reduce GPU compute times and curb the environmental costs of training LLMs~\cite{10.1145/3442188.3445922}.

\section{Background}
\label{sec: background}
% Large language models (LLMs) have emerged as a groundbreaking innovation in the realm of natural language processing. LLMs are pre-trained on a large corpus of data and learn to generate text for various tasks. In the code auto-completion area, various LLMs have been trained with a huge amount of code and show impressive results in the tasks of code generation and code completion. However, the codes in the training dataset are collected and processed in a natural language text data way. Consequently, there are spaces for the 

% introduction

Code compilation plays a pivotal role in code analysis, particularly in the domain of high-performance computing (HPC). It stands to reason that the proportion of compilable code within a training dataset would influence the ability of Large Language Models to generate valid, compilable code. In this section, we explore the significance of compilable code and provide an overview of the existing code datasets utilized for Large Language Model training.

\subsection{Source Code Analysis}
Source code analysis, especially static code analysis, has a rich history that traces its roots back to the early days of programming. Its primary objective has always been to analyze the source code of a program without executing it, enabling developers to detect issues, vulnerabilities, or even adherence to coding standards.

Historically, manual code reviews served as the primary means of analyzing code. However, as software projects grew in complexity and size, manual reviews became increasingly challenging. This precipitated the need for automated code analysis tools.

Over the years, various tools have been developed to cater to different programming languages and purposes. For instance, tools like \textit{Lint} for C, \textit{SonarQube} for multiple languages, and \textit{FindBugs} for Java have become mainstays in the developer's toolbox. These tools automatically inspect the codebase for potential issues, ranging from style violations and potential bugs to more severe security vulnerabilities.

Compilable code plays a crucial role in this process. Only when the code can be successfully compiled can certain types of analyses, especially those that rely on understanding the interplay between different parts of a program, be conducted effectively. Compilable code, essentially, serves as a foundational requirement for many code analysis tools to function optimally. It also enables possibilities like dynamic analysis, where the program's behavior is observed during its execution, and facilitates a more comprehensive understanding of the software's potential behavior in real-world scenarios. For example, Intermediate Representations (IR) of code can only be generated from compilable code by LLVM \cite{10.5555/977395.977673}.

In the context of Large Language Models and their training on code datasets, the presence of compilable code is indispensable. It ensures that the generated code snippets are not just syntactically correct but also semantically meaningful, allowing for a richer and more accurate analysis.

\subsection{Large Language Models and Code Generation}

Large Language Models (LLMs) have emerged as a groundbreaking innovation in the realm of natural language processing. Rooted in deep learning architectures, especially transformer-based \cite{NIPS2017_3f5ee243} models like BERT \cite{devlin2019bert}, GPT \cite{openai2023gpt4}, PaLM \cite{chowdhery2022palm}, and their successors, LLMs have the capability to comprehend and generate human-like text across a myriad of tasks. They have performed exceptionally well on code completion benchmarks like HumanEval \cite{chen2021evaluating} which look at the functional correctness of the code and MBPP \cite{austin2021program}.

The sheer size and scale of LLMs, often trained on vast amounts of diverse textual data, enable them to encapsulate nuanced patterns and intricacies of language. While they were initially oriented towards natural language tasks, their prowess was soon recognized in adjacent domains, most notably in code generation and comprehension.

\subsubsection{LLMs for Code Generation}

The idea of automating code writing isn't novel. However, the introduction of LLMs into this space has revolutionized what is possible. By training LLMs on vast code repositories, these models have demonstrated an ability to understand programming constructs, idioms, and even best practices across multiple languages. For instance, GitHub CoPilot\footnote{\url{https://github.com/features/copilot}}, an AI based pair programming system can generate code directly from natural language problem description. CoPilot is powered by Codex~\cite{chen2021codex}, which is developed by OpenAI and obtained by finetuning 12B parameter GPT models billions of lines of source code.

Platforms like GitHub, StackOverflow, and various code documentation sites have been invaluable resources for training data. When presented with a coding problem or a partial code snippet, LLMs can suggest completions, fix bugs, translate languages~\cite{roziere2020transcoder}, or even generate entire routines~\cite{chen2021codex}. Such capabilities have not only expedited the coding process but have also served as educational tools, assisting novice developers in understanding coding best practices.
Within the research community and industry, several efforts have emerged aiming to curate, refine, and utilize high-quality datasets for training LLMs for code generation tasks. A few notable examples include:

\paragraph{StarCoder:}
StarCoder \cite{li2023starcoder} is a 15B parameter model trained for code generation or completion. The training dataset, the Stack \cite{Kocetkov2022TheStack}, has 1 trillion tokens sourced from a large collection of permissively licensed GitHub repositories.

\paragraph{Code Llama:}
Code Llama \cite{rozière2023code} is an LLM capable of generating code from natural language prompts. It is a code specialized version of Llama2 \cite{touvron2023llama} developed by Meta. It was derived from Llama 2 by training it with code-specific datasets. Code Llama was released in sizes of 7B, 13B and 34B parameters. Each was trained with 500B tokens of code data.

\paragraph{WizardCoder:}
Most code LLMs like StarCoder have been trained on raw code data without instruction fine tuning. WizardCoder\cite{luo2023wizardcoder} with its complex instruction fine tuning by adapting the Evol-Instruct methods for coding tasks has shown to have improved performance for code generation. 

These examples underscore the diversity of approaches and priorities when curating datasets for LLM training in code generation. Each dataset, with its unique strengths and focus areas, contributes to the broader goal of creating LLMs that are proficient, reliable, and versatile in generating code.

However, as highlighted previously, the quality of the training data is paramount. For LLMs to generate effective, efficient, and, most importantly, compilable code, they must be trained on high-quality, error-free codebases. This underscores the importance of curating and refining code datasets for LLM training, ensuring that the resulting models are both accurate and useful in real-world coding scenarios.
\subsubsection{Instruction Fine Tuning}
Instruction fine-tuning on code-specific LLMs typically requires training the model with code-related data, such as code snippets, documentation, or domain-specific programming languages. The goal is to enhance the model's understanding of programming concepts, idioms, and best practices. This approach has the potential to revolutionize software development by automating coding tasks, offering code recommendations, and aiding developers in writing efficient and accurate code. By refining the model's parameters and enhancing its knowledge, instruction fine-tuning enhances its performance in specialized contexts, making it a valuable tool for tailoring state-of-the-art language models to address specific research objectives and real-world challenges. Models like FLAN-T5 \cite{chung2022scaling} have shown the improved performance of incorporating fine tuning. Even models like Code Llama have instruct variants which are instruction fined tuned for different objectives.
\subsubsection{Quality of training dataset for LLMs}

The efficacy of LLMs in the realm of natural language processing has been widely acknowledged. Their adeptness at understanding context, retaining long-term dependencies, and generating coherent text makes them exceptionally well-suited for various NLP tasks. As a result of this success, there has been a growing interest in harnessing the capabilities of LLMs for specialized tasks, particularly in the domain of code generation.

For LLMs to excel in code generation, the choice and quality of the training dataset~\cite{gunasekar2023textbooks} become paramount. Unlike traditional NLP tasks that rely on textual data from books, articles, and websites, code generation requires a different kind of linguistic understanding — one rooted in the logic, structure, and semantics of programming languages.

Consequently, platforms that host vast repositories of code, such as GitHub and StackOverflow, have become pivotal in curating datasets for training LLMs in this domain. The diversity and richness of code available on these platforms, spanning multiple programming languages and tackling myriad computational problems, provide an ideal foundation.

However, merely having access to vast amounts of code is not sufficient. The code needs to be of high quality, devoid of errors, and representative of good programming practices. Additionally, as we have emphasized earlier, the compilability of the code in the dataset is a significant factor. Training on non-compilable or poorly written code can inadvertently teach the model to reproduce such mistakes~\cite{hasabnis2022codequality}, detracting from the model's utility in real-world code generation scenarios causing performance degradation and thus adding noise~\cite{10.1145/3510003.3510160}. Providing high quality code datasets~\cite{gunasekar2023textbooks} to models with lower number of parameters has shown to have comparative performance against models with higher number of parameters with lower quality code datasets.

Thus, meticulous curation, preprocessing, and validation of code datasets are essential to ensure that LLMs trained for code generation are not only proficient but also reliable in generating viable, efficient, and compilable code snippets.

\section{Code Data Validation}
\label{sec: valid}
In this section, we delve into the validation phase of CompCodeVet. Our exploration commences with an evaluation of open-source code datasets for LLM training. Subsequently, we detail our methodology for collecting data tailored for the fine-tuning of an LLM, aimed at effective programming language classification.

\subsection{Code Data Source}
The dataset used in this work mostly comes from two open-source datasets: Stack and HPCorpus~\cite{kadosh2023quantifying}. The Stack~\cite{Kocetkov2022TheStack}, a 6.4 TB dataset of permissively licensed source code in 384 programming languages,
included 54 GB of GitHub issues and repository-level metadata in the v1.2 version of the dataset. 

\subsection{Compilability Evaluation}
\label{sec: compilation_eval}
To assess the compilability of each dataset, we selected random samples comprising 50k C code snippets and 50k C++ code snippets. For the compilation process, we employed the GNU Compiler Collection 12.3 (GCC-12.3). As per Definition 3.1, we opted not to account for linker errors in our analysis.

\begin{definition}
Compilation entails the conversion of source code into machine code or an intermediary representation. This process involves scrutinizing the code for various errors and subsequently generating the pertinent object files. A C or C++ code snippet is deemed ``compilable" if the GCC compilation process concludes without returning any errors. 
\end{definition}

%TODO: I think we should clearly say how we are compiling these programs, given that they could be dependant on others.

\begin{figure}[t]
  \begin{center}
     \includegraphics[width=0.47\textwidth]{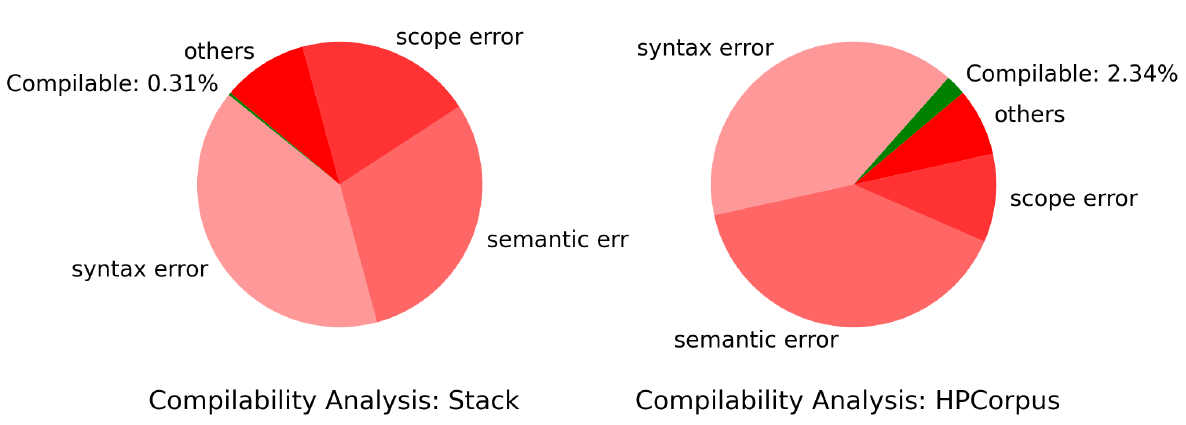}
     \caption{GCC compilation analysis results. The potion of non-compilation slides indicates the top categories.}
    \label{fig: gcc-res}
  \end{center}
  % \vspace{-5mm}
\end{figure}

The results derived from our GCC compilation analysis are detailed in Figure \ref{fig: gcc-res}. The compilability rates for the Stack and HPCorpus datasets are 0.31\% and 2.34\%, respectively. Additionally, a manual investigation of the compiler's report highlighted the predominant causes hindering the compilation of C/C++ code, as follows:
\begin{itemize}
    \item syntax error: These are mistakes in the code structure, such as missing semicolons, parentheses, braces, or incorrect variable names. Syntax errors prevent the compiler from understanding the code.
    \item semantic error: errors such as undefined symbols and type errors.
    \item scope error: using variables outside their scope.
\end{itemize}

\subsection{Programming Label Classification}
The programming language (PL) of a code snippet not only dictates the compiler choice for compilation tests but also influences downstream tasks where the programming language is a significant factor. Through a meticulous inspection of compiler outputs, we discerned that the Stack dataset contains 34.7\% of C code which is mislabeled. Specifically, out of the 50k samples tested from the Stack dataset, 17,350 samples of Objective-C are incorrectly labeled as C code.

Although the programming language of most code snippets is typically inferred from the file extension, there are scenarios in which this approach can lead to mislabeling:
\begin{itemize}
    \item \textbf{Shared File Extensions:} Certain programming languages use identical file extensions. For instance, both C and Objective-C use the ``*.h" extension for header files. However, their syntax and grammatical structures are vastly different. Incorporating mislabeled Objective-C code in the training dataset can adversely impact the model's performance.
    \item \textbf{Code Snippets from Websites:} Websites, such as Stack Overflow, often present code snippets without explicitly mentioning the file extension, making it challenging to automatically identify the correct programming language.

% \item \textbf{Original File Errors:} Many training datasets rely on original file labels to tag the programming language of code snippets. However, there can be instances where the source files themselves carry incorrect labels.

\end{itemize}

\subsection{LLM Fine-tuning for Programming Language Classification}

For the successful invocation of appropriate compilers in CompCodeVet, it's pivotal to accurately classify the programming language of given code snippets. To this end, we fine-tuned the Llama2-7b model specifically for the task of language classification.

\begin{figure}[t]
  \begin{center}
     \includegraphics[width=0.25\textwidth]{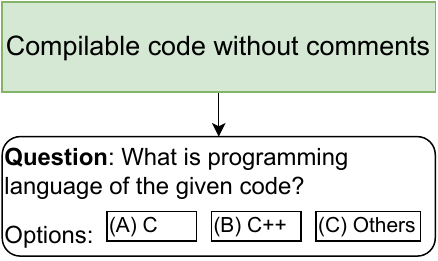}
     \caption{Example of processed data for fine-tuning the model for programming language classification.}
    \label{fig: PL-prompt}
  \end{center}
  % \vspace{-5mm}
\end{figure}

\textbf{Dataset Construction.} We selected 50k compilable code snippets from the HPCorpus \cite{kadosh2023quantifying} dataset, using their associated PL labels as ground truth for the fine-tuning process. In our preprocessing steps, depicted in Figure \ref{fig: PL-prompt}, comments from the code snippets were systematically removed. Furthermore, for the creation of instructive data, we explicitly tagged each snippet with its respective PL label.

\section{Code Data Enhancement}
\label{sec: enhance}
% This section first elucidates our criteria for assessing the quality of a code dataset, specifically focusing on its compilability. Subsequently, we delve into the two steps of CompCodeVet for code validation and enhancement.

% Section\ref{sec: background} elucidates our criteria for assessing the quality of a code dataset, specifically focusing on its compilability. This section elucidates the details of CompCodeVet with respect to code validation and enhancement. 

% LLM-based zero-shot approach to inspect and enhance the quality of training datasets for Code Large Language Models. We characterize the quality of code data based on two primary criteria:

This section outlines the specifics of the dataset enhancement phase of CompCodeVet. In this phase, CompCodeVet uses a compiler-driven chain-of-thought strategy to turn non-compilable code into its compilable counterpart. As depicted in Figure \ref{fig: intro}, CompCodeVet harnesses the capabilities of compilers to steer the reasoning of LLMs, moving away from the conventional approach of relying on a larger-sized LLM.

\subsection{Compiler-guided Chain-of-Thought Prompting}

\citet{wei2022chain} introduced the concept of CoT prompting for multi-step reasoning tasks. In contrast to directly prompting an LLM to rectify a non-compilable code snippet in one go, CompCodeVet systematically addresses each error reported by the compiler, one at a time. Once all errors from the initial compiler output have been addressed, CompCodeVet then compiles the modified code to ensure its validity. If no new errors arise, the code is deemed compilable. If new errors are detected, the system repeats the iterative error-fixing process.

\begin{figure}[t]
  \begin{center}
     \includegraphics[width=0.5\textwidth]{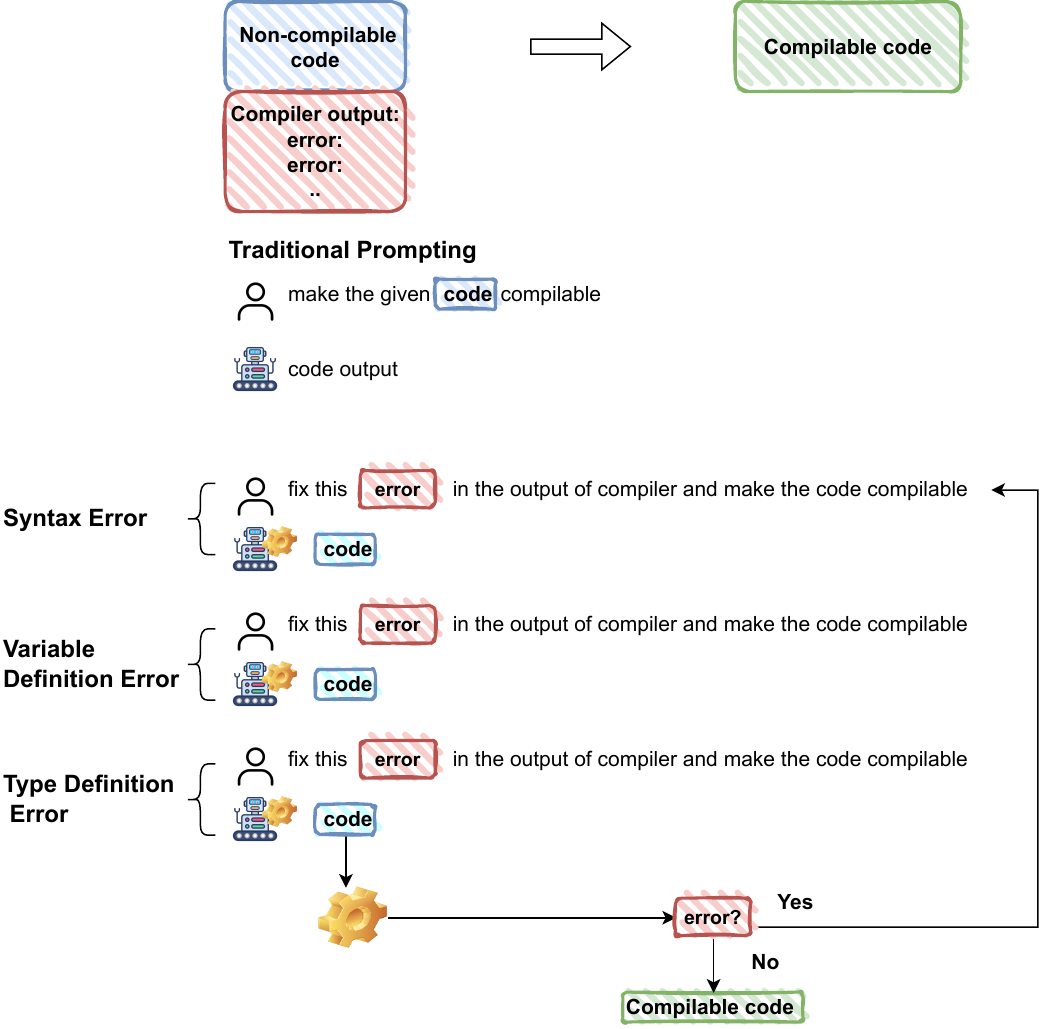}
     \caption{An illustration of compiler-guided chain-of-thought prompting in CompCodeVet.}
    \label{fig: cot}
  \end{center}
  % \vspace{-5mm}
\end{figure}

Although the steps in CompCodeVet's CoT framework can vary based on the specifics of the code and errors encountered, the principle remains consistent: address one error at a time. Rather than requesting the LLM to generate a universally compilable code, the focus is on rectifying individual, identified errors as shown in Figure \ref{fig: cot}. The prompts are structured as:

\begin{graybox}
$P_0$[Given code $C_0$], please rectify [error $e_0$] identified in the compiler output.
\end{graybox}

$P_0$ represents the initial prompt. This step can be mathematically represented as $C_1 = \text{argmax}p(e_0|C_0)$, wherein $C_1$ is the updated code with error $e_0$ addressed. As long as there remain errors to be fixed, subsequent prompts are generated:

\begin{graybox}
$P_1$[Given code $C_1$], please rectify [error $e_1$] indicated by the compiler.
\end{graybox}

Once the current error list is exhausted, the newly generated code undergoes compilation to verify its integrity. If additional errors are discovered, the described iterative process continues. However, to ensure that the process doesn't run indefinitely, we've imposed a maximum iteration limit of $K$. This constrains the number of iterations within CompCodeVet to not exceed $K$.

Once the current error list is exhausted, the newly generated code undergoes compilation to verify its integrity. If additional errors are discovered, the described iterative process continues until the code is fully rectified and compiled without errors.

 To assure the process does not run infinitely, we set a maximum iteration limit $K$ to limit the iteration within CompCodeVet does not go beyond $K$.

% \begin{enumerate}
%     \item \textbf{Programming Language (PL) labels:} Accurate labeling of the programming language for code data is indispensable. It not only aids in selecting the appropriate compilers for static analysis but also ensures the consistency and relevance of the training data.
    
%     \item \textbf{Compilability:} As underscored in the Background section, compilable code is pivotal in the domain of code analysis. Moreover, the quality of training data significantly influences LLM performance. Therefore, a training code dataset with a higher proportion of compilable code is deemed of superior quality.
    
%     % \item \textbf{Criterion:} 
% \end{enumerate}

% As depicted in Figure \ref{fig: pipeline}, we harness the capabilities of both LLMs and compilers to rectify incomplete code, rendering it compilable. Fine-tuning the LLMs necessitates an instructive dataset.

\subsection{The BrokenComp-instruct Dataset}

As previously illustrated, we leverage the combined strengths of LLMs and compilers to correct incomplete code, ensuring its compilability. To refine the LLM's reasoning capabilities at every juncture, it is essential to fine-tune the model, targeting the most prevalent compilation errors as identified in Section \ref{sec: compilation_eval}. To this end, we introduce an instructive dataset, termed BrokenComp-instruct. This dataset is meticulously curated to introduce a single error in each instruction, and it is paired with the corresponding compiler error output for guidance.

% Fine-tuning the LLMs enables pre-trained LLMS to boost their performance in compilable code generation. This process allows us to mold the model's output to adhere to our needs. . 

\begin{figure}[!ht]
  \begin{center}
     \includegraphics[width=0.47\textwidth]{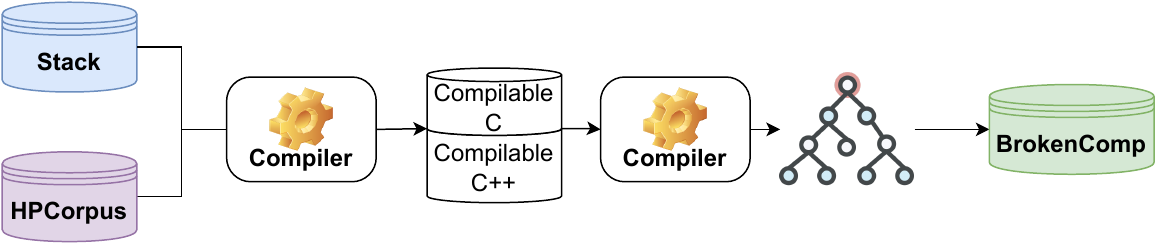}
     \caption{Construction Process of the BrokenComp Dataset.}
    \label{fig: BrokenComp}
  \end{center}
  % \vspace{-5mm}
\end{figure}

\textbf{Dataset collection.} Figure \ref{fig: BrokenComp} illustrates the construction process of the BrokenComp-instruct dataset. To ascertain the accuracy of the target-generated code in the instructions, we initiated our process by collecting compilable C/C++ code. This approach also addresses the concerns highlighted in Section \ref{sec: compilation_eval} regarding the potential mislabeling of code data by datasets. We ended by collecting 50k compilable C code and 50k compilable C++ code. We also filtered out  code with length beyond the common input token limits for popular LLMs.

\textbf{Creating broken code.} Starting with the 100k compilable code snippets, we introduced alterations as shown in Algorithm \ref{alg: create_1error} based on the following operations using the Abstract Syntax Trees (AST):
\begin{itemize}
    \item Constructed the code’s AST and randomly selected variable leaf nodes, subsequently deleting the corresponding variable initialization.
    \item Identified a user-defined type through the AST and eliminated its definition.
    \item Introduced syntax errors by randomly eliminating operators and parentheses.
\end{itemize}

\begin{algorithm}
\caption{Introduce Only One Compilation error using AST}
\label{alg: create_1error}

\begin{algorithmic}[1]
\STATE $AST \gets \text{generateAST}(code)$ \COMMENT{Parse code to generate AST}
\STATE $varNodes \gets \text{findAllVariableLeafNodes}(AST)$

\IF{$\text{length}(varNodes) > 0$}
    \STATE $randomNode \gets \text{selectRandom}(varNodes)$
    \IF{$\text{hasInitialization}(randomNode)$}
        \STATE $\text{deleteInitialization}(randomNode)$
    \ENDIF
\ENDIF

\STATE $typeNodes \gets \text{findAllUserDefinedTypes}(AST)$
\IF{$\text{length}(typeNodes) > 0$}
    \STATE $randomTypeNode \gets \text{selectRandom}(typeNodes)$
    \STATE $\text{deleteTypeDefinition}(randomTypeNode)$
\ENDIF

\STATE $modifiedCode \gets \text{traceASTtoCode}(AST)$ \COMMENT{Convert modified AST back to code}
\STATE \textbf{return} $modifiedCode$
\end{algorithmic}

\end{algorithm}

These steps ensured the creation of a dataset that would challenge and thus refine the LLM's capacity to rectify and compile broken code.

\textbf{Creating fine-tuning instructions.} To optimize the performance of large language models (LLMs) in handling broken code scenarios, we crafted specific instructions paired with the broken code and the associated compiler output. This approach ensures that the LLMs gain a nuanced understanding of how to rectify common errors present in the code.

\begin{graybox}
\# \textbf{Instruction}:

\{"Fix the compiler error of the given {PL} code: {compiler error}"\}

\# \textbf{Input}:

\{broken code\}

\# \textbf{Response}: 

\{original compilable code\}

\end{graybox}

\section{Evaluation}
\label{sec: eval}
In this section, we present and analyze the outcomes derived from our proposed approach, shedding light on its efficacy and implications.

\subsection{Experimental Platform}
Throughout both the code validation and data enhancement stages, we adopted parameter-efficient fine-tuning utilizing the Low-Rank Adaptation (LoRA) \cite{hu2021lora} method, sidestepping the need for comprehensive model fine-tuning.
For the task of programming language classification, we selected the official release of Llama2-7b. Conversely, for generating compilable code, our choice was the official iteration of CodeLlama2-instruct-7b. All model interactions were facilitated using the Huggingface Transformers toolkit \cite{wolf2020huggingfaces}. During the training phase, the models were fine-tuned to align their outputs with the reference responses, leveraging the cross-entropy loss.
We followed the Alpaca LoRA project for hyper-parameter settings. All training experiments share the same software environment and were carried out on 2 Nvidia A100 40GB GPUs.

\subsection{Code Validation Results}
During the code validation phase, our focus was on assessing the efficacy of programming language classification. The prompts for the test mirrored the format used during fine-tuning, as illustrated in Figure \ref{fig: PL-prompt}. We gauged accuracy by comparing the model's output against the ground truth labels.

\textbf{Test dataset.} Our test dataset was curated using Google BigQuery to crawl code from GitHub, encompassing ten widely-used programming languages: C, C++, Python, Objective-C, Assembly, Java, Go, C\#, Ruby, and R. Each language category comprises of 1,000 samples.

\textbf{Test results.} Our model was benchmarked against the out-of-the-box Llama2-7b, GPT3.5-turbo, and GPT-4. We employed OpenAI's API for inference on the last two mentioned models. Across the board shown in Table 1 , all models showcased impressive metrics in terms of precision, recall, and F1 score. This underscores the adeptness of LLMs in discerning patterns from their training data. Notably, CompCodeVet's fine-tuned  Llama2-7b outperformed its pre-trained counterpart. This improvement can be attributed to the model's capability to emphasize language-specific keywords at distinct tokens and its fine-tuning on high-quality data. While GPT-4 delivered top-tier performance, CompCodeVet achieved comparably impressive results with a significantly more compact model (1.76 trillion vs. 7 billion).

Upon examining the misclassified cases, we observed that most of these cases occur between C and C++. These samples posed challenges even for human developers. Such ambiguity can be ascribed to the intricate relationship and shared syntax between the two programming languages. 

\begin{table}[h]
\label{tb:res1}
\centering
\caption{Programming Language Labeling Results}
\begin{tabular}{lccc}
\toprule
Model       & Precision & Recall & F1 \\
\midrule
CompCodeVet & 0.96      & 0.94   & 0.95 \\
Llama2-7b  & 0.93      & 0.92   & 0.92 \\
GPT3.5-turbo       & 0.95      & 0.95   & 0.95 \\
GPT4       & \textbf{0.97}      & \textbf{0.96}   & \textbf{0.96} \\
\bottomrule
\end{tabular}
\end{table}

\textbf{Ablation Study.} Compared to the pre-trained Llama2-7b model, the variant of CompCodeVet was fine-tuned with keywords in programming languages as special tokens. The results in Table 2 suggest the importance of compiler guidance when applying LLMs in code analysis tasks and the importance of high-quality data. We followed the same process to fine-tune the pre-trained Llama2-7b model with and without the special tokens for an ablation study. We also control the training data size in this ablation study.

Table \ref{table:ablation_study} presents a comparison between Llama2-7b models, with and without the addition of special tokens (ST), across different dataset sizes. Our analysis indicates that incorporating special tokens positively influences the precision score, though no discernible impact is observed on the recall score. This might be attributed to the fact that the special tokens don't encompass all the programming language keywords present in the test set, particularly benefiting the identification of C and C++ codes.

As seen from Table \ref{table:ablation_study}, the size of the fine-tuning dataset has a noticeable impact on the model's performance. The results demonstrate a clear correlation between the volume of training data and the model's effectiveness across both precision and F1 scores.

During our manual evaluation, we observed that pre-trained models, when fine-tuned with keywords as special tokens, demonstrated enhanced proficiency in differentiating between C and C++ code with keywords as special tokens.

\begin{table}[!htb]
    \centering
    \caption{Ablation study results comparing Llama2-7b (Llama2) with and without special tokens(ST), fine-tuned with various dataset sizes.}
    \begin{tabular}{|l|l|l|l|l|}
        \hline
        \textbf{Model} & \textbf{Dataset} & \textbf{Precision} & \textbf{Recall} & \textbf{F1} \\
        \hline
        Llama2+ST & 1K & 0.93 & 0.92 & 0.92 \\
        \hline
        Llama2 & 1K & 0.91 & 0.92 & 0.91 \\
        \hline
        Llama2+ST & 10k & 0.92 & 0.91 & 0.91 \\
        \hline
        Llama2 & 10k & 0.91 & 0.91 & 0.91 \\
        \hline
        Llama2+ST & 30k & 0.95 & 0.91 & 0.93 \\
        \hline
        Llama2 & 30k & 0.93 & 0.91 & 0.92 \\
        \hline
        Llama2+ST & 50k & 0.95 & 0.93 & 0.94 \\
        \hline
        Llama2 & 50k & 0.92 & 0.93 & 0.92 \\
        \hline
    \end{tabular}
    
    \label{table:ablation_study}
\end{table}

\subsection{Code Enhancement Results}
The primary objective of the code enhancement phase within CompCodeVet is to transform non-compilable code into a compilable state. Listing~\ref{lst:no_comp_2} shows an output example of CompCodeVet with Listing~\ref{lst:no_comp_1} as input.

\begin{lstlisting}[language=C, caption=Instance of non-compilable code in HPCorpus, label=lst:no_comp_1]
{
    void comm_clean()
    {
        comm_close();
        if (port_name)
            free(port_name);
        port_name = NULL;
    }
}
\end{lstlisting}

\begin{lstlisting}[language=C, caption=Instance of compilable code generated by CompCodeVet, label=lst:no_comp_2]
#include <stdlib.h> 
char *port_name = NULL; 
void comm_clean()
{
    comm_close();
    if (port_name)
        free(port_name);
    port_name = NULL;
}
\end{lstlisting}

\textbf{Test dataset.} Our test dataset was curated by amassing 5k C and 5k C++ code snippets during the development of the BrokenComp Dataset, as depicted in Figure \ref{fig: BrokenComp}. For efficient inference with CompCodeVet, we capped the iteration limit, $K$, at three. 

\textbf{Test results.} We compared the performance of CompCodeVet with the out-of-the-box CodeLlama-7b-instruct, starchat-alpha, and WizardCoder-15b. All the models are open-source LLMs specifically pre-trained with code data. The tests were performed with the same prompts in Figure \ref{fig: cot}.

\begin{table}[h]
\centering
\caption{Comparison of Model Compilability(Comp) between CompCodeVet, CodeLlama-7b-instruct(CodeLlama), StarChat Alpha(StarChat), and WizardCoder-15B(WizardCoder)}
\label{tb:res2}
\begin{tabular}{p{2.5cm}cc}
\toprule
Model                & Comp-C (\%) & Comp-C++ (\%) \\
\midrule
CompCodeVet          & 10.6                 & 8.4 \\
CodeLlama & 2.1                  & 1.5 \\
StarChat       & 0.2                  & 0.1 \\
WizardCoder     & 0.1                  & 0.1 \\
\bottomrule
\end{tabular}
\end{table}

\textbf{Ablation Study.} CompCodeVet refines the code in an iterative manner, drawing insights from the compiler's feedback. This iterative approach, however, poses a risk of infinite loops if the LLM within the CoT framework either fails to rectify existing errors or inadvertently introduces new ones. This scenario resembles signal amplification, albeit with escalating errors instead of signals. In this context, setting an appropriate iteration limit of $K$ becomes paramount. Table \ref{tb:iterations} reveals that $K=4$ yields the most favorable results on a compact test set consisting of 1k samples. However, a setting of $K=3$ provides outcomes that are nearly as optimal. Given considerations of computational efficiency, we opted for $K=3$ for the experiments presented in Table 3.

\begin{table}[h]
\centering
\caption{Compilability (Comp) of Code at Different Iterations (K)}
\label{tb:iterations}
\begin{tabular}{c|cc}
\toprule
K & Comp-C (\%) & Comp-C++ (\%) \\
\midrule
1 & 3.4 & 2.7 \\
2 & 6.8 & 3.1 \\
3 & 9.4 & 7.9 \\
4 & 9.4 & 8.1 \\
5 & 9.4 & 8.1 \\
\bottomrule
\end{tabular}
\end{table}

\section{Related Works}
\label{sec: relwork}

Standard code datasets~\cite{kocetkov2022stack,li2022competition} suffer from several drawbacks, such as incompleteness, inaccurate information, and ambiguity (e.g., incomplete snippets, incorrect labels, non-compilable), which would impact the quality of trained machine learning models for code. The field of code cleaning and data preprocessing has seen significant research and development efforts over the years. The research on using LLMs for code cleaning includes efforts to transform natural language descriptions into data cleaning code. These endeavors have shown promising results in simplifying code cleaning tasks. CodeBERT \cite{feng2020codebert} is a pre-trained model that can understand and generate code from natural language comments, thus streamlining data preprocessing through human-readable descriptions. In recent years, a successful approach to improve model performance has been to scale up the model parameters and training data. \cite{gunasekar2023textbooks} has clearly shown us that improving the quality of the training dataset while using lesser number of parameters and fewer tokens can have on par performance with state-of-the-art LLMs which use trillions of parameters and tokens. Alternately, the field of software engineering has extensive set of tools to measure code quality, complexity, among other metrics~\cite{ludwig2017smc, mccabe1976complexity}, and Hasabnis et al.~\cite{hasabnis2022codequality} has recently employed these tools to evaluate quality of open-source code repositories on GitHub~\cite{hasabnis2022gitrank}.

\section{Conclusion}
\label{sec: conclusion}
In this work, we presented CompCodeVet, a novel compiler-guided approach to leverage Large Language Models (LLMs) for the tasks of code validation and enhancement. Our main contribution lies in employing the power of compilers as teachers, steering away from the conventional reliance on larger LLMs, and emphasizing the importance of multi-step reasoning in generating and validating code.

Our findings underscore the inefficiencies in current LLMs when faced with complex reasoning tasks, particularly in the context of code generation and comprehension. CompCodeVet's unique approach to employing a chain of thought (CoT) with the compiler's feedback loop showed promise in addressing such challenges, emphasizing the benefit of stepwise error resolution.

\bibliography{main}
\bibliographystyle{mlsys2024}

%%%%%%%%%%%%%%%%%%%%%%%%%%%%%%%%%%%%%%%%%%%%%%%%%%%%%%%%%%%%%%%%%%%%%%%%%%%%%%%
%%%%%%%%%%%%%%%%%%%%%%%%%%%%%%%%%%%%%%%%%%%%%%%%%%%%%%%%%%%%%%%%%%%%%%%%%%%%%%%
% SUPPLEMENTAL CONTENT AS APPENDIX AFTER REFERENCES
%%%%%%%%%%%%%%%%%%%%%%%%%%%%%%%%%%%%%%%%%%%%%%%%%%%%%%%%%%%%%%%%%%%%%%%%%%%%%%%
%%%%%%%%%%%%%%%%%%%%%%%%%%%%%%%%%%%%%%%%%%%%%%%%%%%%%%%%%%%%%%%%%%%%%%%%%%%%%%%
%\appendix
%\section{Please add supplemental material as appendix here}
%
%Put anything that you might normally include after the references as an appendix here, {\it not in a separate supplementary file}. Upload your final camera-ready as a single pdf, including all appendices.

%%%%%%%%%%%%%%%%%%%%%%%%%%%%%%%%%%%%%%%%%%%%%%%%%%%%%%%%%%%%%%%%%%%%%%%%%%%%%%%
%%%%%%%%%%%%%%%%%%%%%%%%%%%%%%%%%%%%%%%%%%%%%%%%%%%%%%%%%%%%%%%%%%%%%%%%%%%%%%%

\end{document}